\documentclass[conference]{IEEEtran}
\IEEEoverridecommandlockouts
\usepackage{cite}
\usepackage{amsmath,amssymb,amsfonts}
\usepackage{algorithmic}
\usepackage{graphicx}
\usepackage{textcomp}
\usepackage{xcolor}
\usepackage[export]{adjustbox}
\usepackage{booktabs,makecell}
\usepackage{bm}

\def\BibTeX{{\rm B\kern-.05em{\sc i\kern-.025em b}\kern-.08em
    T\kern-.1667em\lower.7ex\hbox{E}\kern-.125emX}}
\begin{document}

\title{
\huge
Octopus-inspired Distributed Control for Soft Robotic Arms: A Graph Neural Network–Based Attention Policy with Environmental Interaction
\vspace{-0.7em}
}

\author{
\IEEEauthorblockN{
Linxin Hou\IEEEauthorrefmark{1}, 
Qirui Wu, 
Zhihang Qin, 
Yongxin Guo\IEEEauthorrefmark{1}, 
Cecilia Laschi\IEEEauthorrefmark{1}
}
\thanks{This work was supported by Start-up grant RoboLife (Soft Robots with morphological adaptation and life-like abilities), DESTRO (Dextrous, strong yet soft robots), ITALY–SINGAPORE grant, MAE (Italy) and A*STAR (Singapore), Grant \#R22I0IR124, and Bridging fund (AI-Driven Soft Robots for Marine and Unstructured Environments).

L. Hou is with Department of Electrical and Computer Engineering, National University of Singapore, Singapore (e-mail: hou.linxin@u.nus.edu)

Q. Wu is with Department of Computer Science, National University of Singapore, Singapore

Z. Qin is with Department of Mechanical Engineering, National University of Singapore, Singapore

Y. Guo is with Department of Electrical and Computer Engineering, National University of Singapore, Singapore; Department of Electrical Engineering, City University of Hong Kong, HKSAR, China

C. Laschi is with the Advanced Robotic Centre, National University of Singapore, Singapore; Department of Mechanical Engineering, National University of Singapore, Singapore.
}
\vspace{-2.5em}
}

\def\IEEEtitletopspaceextra{0.2in}
\maketitle

\begin{abstract}
This paper proposes SoftGM, an octopus-inspired distributed control architecture for segmented soft robotic arms that learn to reach targets in contact-rich environments using online obstacle discovery without assuming full obstacle geometry at the start of each episode. SoftGM formulates each arm section as a cooperative agent and represents the arm-environment interaction as a graph. SoftGM uses a two-stage graph attention message passing scheme following a Centralised Training Decentralised Execution (CTDE) paradigm with a centralised critic and decentralised actor. We evaluate SoftGM in a Cosserat-rod simulator (PyElastica) across three tasks that increase the complexity of the environment: obstacle-free, structured obstacles, and a wall-with-hole scenario. Compared with six widely used MARL baselines (IDDPG, IPPO, ISAC, MADDPG, MAPPO, MASAC) under identical information content and training conditions, SoftGM matches strong CTDE methods in simpler settings and achieves the best performance in the wall-with-hole task. Robustness tests with observation noise, single-section actuation failure, and transient disturbances show that SoftGM maintains comparable performance under non-ideal simulated conditions while keeping control effort bounded, suggesting that selective contact-relevant information routing improves resilience in the tested settings.

\end{abstract}

\section{Introduction}
\begin{figure*}[!t]
    \centering
    \includegraphics[width=\textwidth, keepaspectratio]{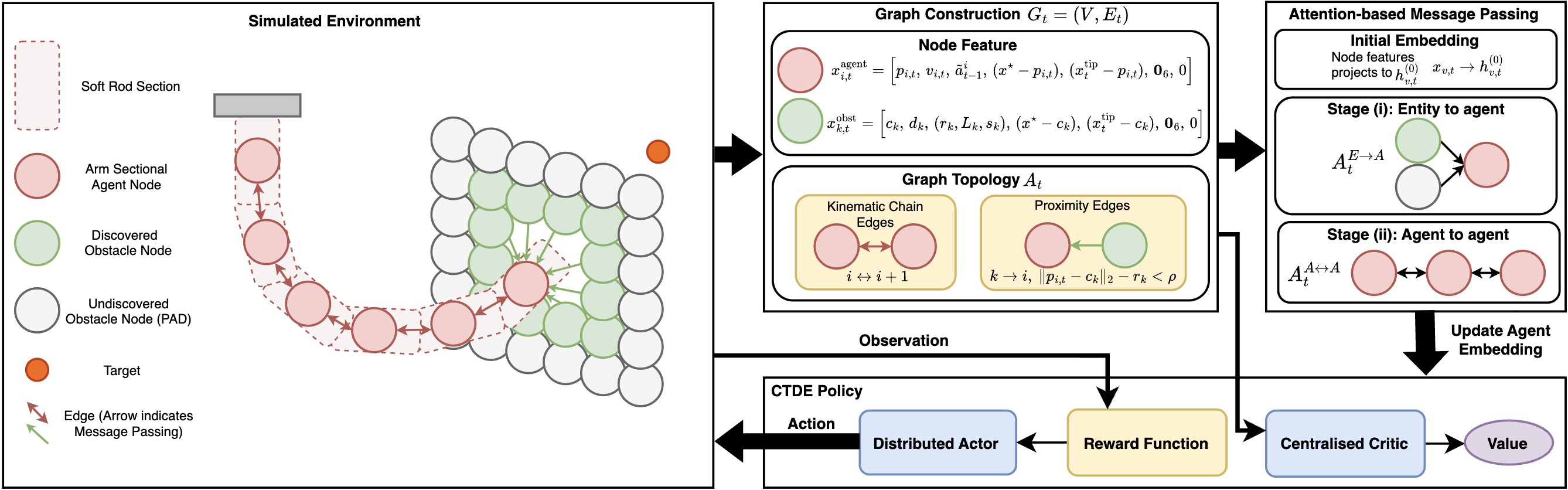}
    \caption{Overview of the SoftGM framework with graph construction and attention-based message passing.}
    \label{fig1} 
    \vspace{-1em} 
\end{figure*}

Soft robotic arms are often idealised as continuum bodies, where deformation is described by continuous state fields \cite{b1, b2, b3}. In practice, tractable modelling and control rely on piecewise or multi-segment approximations of this continuum description, reducing the infinite-dimensional morphology to a finite set of segment states \cite{b4}. Although finite-dimensional, the resulting system remains strongly coupled, such that local actuation or contact can influence neighbouring segments and propagate through the entire body \cite{b5, b6, b7}. Consequently, achieving accurate dynamics under real-time constraints is challenging. Coupling increases computational demands, and external contact introduces non-smooth, state-dependent forces, leading to an inherent trade-off between fidelity and tractability.

The octopus' nervous system represents one of the most promising frontiers in soft robotics, offering bio-inspired solutions to the soft robotic control challenges. The majority of octopus neurones are distributed in their arms rather than centralised in the brain. Their high-dimensional arm behaviours are coordinated through distributed sensing and local reflexes rather than a single centralised controller \cite{b8}. Neurobiological and robotics-oriented analyses show that such distributed architectures help manage extremely high-dimensional limb control by focusing on local sensing, local actuation, and local reflex-like behaviours that coordinate through neighbour communication \cite{b8, b9, b10}. Translating this insight into segmented soft robots, each segment can compute a local action or local policy output using its own observations and neighbouring messages, reducing the burden on any single centralised module and improving robustness under partial observability \cite{b11, b12, b13, b14}. Inter-agent message passing mirrors neural signalling among arm ganglia, while
the central processor plays the role of the integrative brain, enabling coordinated behaviour.

In complex environments, control extends beyond nominal geometric trajectory tracking. In nature, the octopus uses arms to explore the environment by performing actions such as probing, sweeping, and conforming. These actions create informative contacts, which re-shape subsequent motions, using rich peripheral feedback from the arm and suckers \cite{b8,b15,b16}. This biological strategy motivates our emphasis on an online environmental exploration for segmented soft robotic arms. Since contact is local, its implications must be integrated across neighbouring segments through the interaction pathways of the body \cite{b17}. Previous work has shown that physical interaction can support informative environmental inference, but complexity increases rapidly with robot length, sensing density, and the number of simultaneous contacts \cite{b18,b19}. Many existing approaches rely on centralised representations and fixed-dimensional inputs, which can lead to information bottlenecks \cite{b4,b5,b20}. Moreover, under intermittent contact with multiple obstacles, an online controller must decide which contacts are most relevant at any given time.

Therefore, this paper proposes SoftGM, an octopus-inspired distributed Multi-agent Reinforcement Learning (MARL) control architecture using online environmental interaction. The proposed method is based on a Graph Neural Network (GNN) which aligns with the intrinsic structure of segmented soft bodies and the environment \cite{b21}. We represent the soft arm as a graph whose nodes correspond to continuum segments and edges encode physical adjacency and potential interactions between segments. Node features encode local proprioceptive and tactile observations, while contact events are incorporated as additional interaction features in the graph. Building on the message-passing mechanism, the proposed architecture added Graph Attention Networks (GAT) to provide learnable attention weights that prioritise the most relevant neighbours \cite{b22}. In complex environments, this mechanism enables the controller to focus on the dominant nodes at a given time while suppressing irrelevant information. The overall design of the SoftGM framework is shown in Figure \ref{fig1}.

The main contributions of SoftGM are:
\begin{itemize}
    \item A bio-inspired MARL distributed control formulation for segmented soft arms that supports contact-centric operation and environmental interaction.  
    \item A graph-based control architecture that preserves physical topology and local interactions while maintaining global consistency through the passing of graph messages.
    \item A two-stage attention-based message-passing mechanism that adaptively prioritises discovered contact-relevant entities and neighbouring section information in simulated cluttered environments.
\end{itemize}

\section{Related Work}
\subsection{Soft Robotic Arm Control}
Previous work on soft/continuum robotic arm control spans model-based and learning-based approaches. Model-based controllers typically rely on reduced-order kinematics/dynamics coupled with robust designs. Although these methods offer interpretability and stability, their computational and modelling burdens grow rapidly with high-dimensional dynamics, contact, and self-collision, motivating reduced-order modelling alternatives. \cite{b4, b6, b7, b23, b24}.

Learning-based methods such as supervised models, Koopman learning, recurrent model-based Model Predictive Control (MPC), and deep reinforcement learning (RL), provide flexible alternatives and have achieved closed-loop control and contact-involving manipulation behaviours. However, they consistently expose challenges in data efficiency, robustness and safety under distribution shift and sim-to-real transfer, particularly as the number of controllable sections increases \cite{b5, b20, b25, b26, b27, b28, b29}. Moreover, many learning pipelines still assume a fixed-dimensional global state, making it difficult to transfer across discretization or modular configurations without retraining.

Cluttered manipulation motivates a parallel thread that integrates tactile sensing and contact estimation, including contact detection, tactile sensing, and vision-free perception \cite{b17, b18, b19, b30, b31}. However, a persistent gap is that many controllers remain centralised and fixed-dimensional, which create information bottlenecks and degrade scalability as the number of simultaneous contacts grows \cite{b4, b5, b14, b23}. In changing environments rich in obstacles, contact information is sparse, local, and time-varying, and must be integrated with a globally coupled body for representations that preserve local–global consistency while allowing localised distributed decision making. Recent work has also begun to address scalability via modular multi-agent policies for hyper-redundant continuum robots \cite{b32}. Although the results support the value of attention-enhanced modular control, it focuses primarily on agent-to-agent coordination and does not explicitly model external obstacles as discovered entities that enter and leave the robot’s information pool.

\subsection{Graph Attention in MARL for Conventional Robots}
Graph attention has already been used successfully in conventional robotic tasks where agents must coordinate under partial observability, dynamic interaction graphs, and variable numbers of agents. In multi-robot exploration, MARVEL learns a decentralised policy with graph attention to fuse spatial structure with agent state, improving coordination and generalising across team sizes \cite{b33}. For multi-agent navigation with obstacles, InforMARL builds an agent–entity interaction graph and applies attention-based message passing for actor/critic, showing improved sample efficiency and scaling to arbitrary numbers of agents and obstacles \cite{b34}. In dense crowd navigation, MultiSoc combines attention-based edge selection with GAT aggregation to simplify interactions, allowing scalable coordination between different robot–human ratios and behaviour models \cite{b35}. MAGEC supports relational inductive biases for robustness in cooperative control \cite{b36}. COMPOSER is close to SoftGM in its use of modular policies for elongated robotic bodies, but it primarily models coordination among body modules and does not explicitly introduce discovered environmental entities as persistent graph nodes \cite{b32}. InforMARL is close in its agent-entity graph formulation, but it targets conventional multi-agent navigation rather than ordered continuum-body control with local contact propagation. SoftGM combines these directions by using a shared modular actor over arm sections while explicitly routing discovered obstacle information through a topology-aware soft-arm graph.

\vspace{-0.5em}
\section{Methodology}
\subsection{Distributed Control Policy}
SoftGM formulates a soft robotic arm control as a Decentralised Partially Observable Markov Decision Process (Dec-POMDP) with $N$ agents \cite{b37}, where each agent corresponds to one actuator control point. The underlying state of the simulator at time $t$ is $s_t\in\mathcal{S}$, where $S$ includes the complete state of the Cosserat rod and any fixed obstacle geometry. The joint action of the arm is $a_t=(a_t^1,\dots,a_t^N)$. The agent $i$ outputs a continuous action $a_t^i\in[-1,1]^{d_a}$, where $d_a=2$ (normal and binormal). The simulator applies a physical torque vector
$
u_t^i=\tau_{\max}\odot a_t^i,
$
where $\tau_{\max}\in\mathbb{R}^{d_a}_{>0}$ denotes per-axis torque limits and $\odot$ is the element-wise product.

Each agent controls a local B-spline coefficient along the rod. Denoting the normalised arc-length by $s\in[0,1]$ and the basis of the B-spline by $b_i(s)$, the couple-torque coefficient for the director $k$ is scaled from the normalised action as
$\alpha_{i,t}^{(k)}=\tau_{\max}^{(k)}\,a_{i,t}^{(k)}$ \cite{b38},
The resulting distributed torque field can be written in compact form as
$\tau_t^{(k)}(s)=\sum_{i=1}^{N} b_i(s)\,\alpha_{i,t}^{(k)}$.
In the MARL implementation, the actor network is shared among agents using parameter sharing, and produces $N$ actions in one forward pass from a shared graph observation. All agents receive the same team reward $r_t=r(s_t,a_t)$, and the goal is to maximise the discounted return
$J(\pi)=\mathbb{E}_{\pi,P}\big[\sum_{t=0}^{T-1}\gamma^t r_t\big]$.

SoftGM adopts the CTDE paradigm \cite{b39}. During execution, each agent acts using a decentralised policy, while during training, a centralised value function is learnt to reduce variance. Specifically, the actor is a squashed Gaussian policy. Given an agent embedding $g_{i,t}$ from the GNN, the policy mean is
$\mu_{i,t}=f_\pi\!\big([o_t^i \,\|\, g_{i,t}]\big)$,
then $z_{i,t}\sim\mathcal{N}(\mu_{i,t},\mathrm{diag}(\sigma^2))$, where $\sigma$ denotes the action standard deviation, and the final action is the bounded output $a_t^i=\tanh(z_{i,t})$. The critic views the full graph with all nodes and edges and pools only over agent-node embeddings to obtain a single state summary,
$\bar g_t=\frac{1}{N}\sum_{i=1}^{N} g_{i,t}$, then predicts the state value as $V_\phi(s_t)\approx f_V(\bar g_t)$. The critic uses global information during training, while the actor remains distributed at execution.

The end-effector (arm tip) position is defined as $x_t^{\mathrm{tip}}$ and the goal is defined as $x^\star$. The distance between them is thus $d_t=\|x_t^{\mathrm{tip}}-x^\star\|_2$ and the one-step progress is $\Delta d_t=d_{t-1}-d_t$.
The total reward used by the training wrapper is
\begin{align*}
r_t
&=
\underbrace{(-d_t)}_{\text{base shaping}}
+
\underbrace{\lambda_p\,\Delta d_t}_{\text{progress bonus}}
-
\underbrace{\lambda_{\Delta a}\,\frac{1}{Nd_a}\sum_{i=1}^N \|a_t^i-a_{t-1}^i\|_2^2}_{\text{action smoothness}} \\
& -
\underbrace{\lambda_t}_{\text{time}}
+
\underbrace{\lambda_c\,\mathbb{I}[C_t>0]+\lambda_{c,n}\,C_t}_{\text{collision shaping}}
+
\underbrace{\lambda_d\,\min\{N_t^{\mathrm{new}},K_{\max}\}}_{\text{discovery bonus}}\\
& -
\underbrace{\lambda_s\,\mathbb{I}[C_t>0]\mathbb{I}[|\Delta d_t|<\varepsilon]}_{\text{stuck penalty}}
+
\underbrace{\lambda_{\mathrm{succ}}\,\mathbb{I}[d_t<r_{\mathrm{succ}}]}_{\text{success bonus}}.
\label{eq:total_reward}
\end{align*}
Here, $C_t$ is the number of rod nodes in contact with obstacles computed geometrically, $N_t^{\mathrm{new}}$ is the number of newly discovered obstacle segments at time $t$, and $K_{\max}$ caps how many discoveries can be rewarded per step. All terms in the reward function are scalar and are broadcast to every agent, making the task fully cooperative. The reward combines dense goal-reaching guidance, efficient control, and contact-driven exploration. The distance and progress terms guide the arm toward the target, the time and action-smoothness penalties discourage slow or oscillatory behaviours, the discovery bonus encourages informative contact with previously unknown obstacle segments, and the stuck penalty discourages prolonged contact without progress. Thus, obstacle interaction is not treated purely as avoidance; it is useful when it reveals task-relevant environmental structure.

\subsection{Graph Construction}
At each timestep, the multi-agent observation is converted into a directed graph $G_t=(V,E_t)$ so that the policy can reason jointly about agent coordination, goal-reaching, and local proximity sensing. The algorithm uses a fixed node budget $|V|=N+N_{\mathrm{obs,max}}$.
The first $N$ nodes are the agent nodes, which are always present. The remaining $N_{\mathrm{obs,max}}$ slots are reserved for the nodes of the obstacle-segment entity. Those nodes are treated as PAD nodes before they are discovered. A discrete node type $\tau_v\in\{\text{AGENT},\text{OBST},\text{PAD}\}$ is assigned and it learns a type embedding $e(\tau_v)$ that will be concatenated to node features before message passing.

Each node has a feature vector $x_{v,t}\in\mathbb{R}^{22}$. We use a shared input dimension for all node types and denote by
$z_{\mathrm{pad}}=\mathbf{0}_{7}\in\mathbb{R}^{7}$ the reserved padding channels. For an agent node $i$, the feature contains local kinematics and relative geometry of the goal and tip:
\[
x^{\mathrm{agent}}_{i,t}=
\big[
p_{i,t},\, v_{i,t},\, \tilde a^i_{t-1},\, (x^\star-p_{i,t}),\, (x_t^{\mathrm{tip}}-p_{i,t}),\, z_{\mathrm{pad}}
\big],
\]
where $p_{i,t},v_{i,t}\in\mathbb{R}^3$ are the position and velocity of the rod node assigned to agent $i$, and $\tilde a^i_{t-1}=[0,a^{i,n}_{t-1},a^{i,b}_{t-1}]\in\mathbb{R}^3$ is the previous 2-D action padded with a zero tangential component to match the local material-frame representation. The policy still actuates only the normal and binormal directions, so $d_a=2$. Obstacle geometry is intentionally omitted from agent nodes, so that obstacle awareness must come through discovered entity nodes. For a discovered obstacle-segment node $k$, we encode a local geometric descriptor available in the simulator, including cylinder parameters and relative geometry to the goal and arm tip:
\[
x^{\mathrm{obst}}_{k,t}=
\big[
c_{k},\, d_{k},\, \eta_k,\, (x^\star-c_k),\, (x_t^{\mathrm{tip}}-c_k),\, z_{\mathrm{pad}}
\big],
\]
where $c_k\in\mathbb{R}^3$ is the obstacle segment centre, $d_k\in\mathbb{R}^3$ is the obstacle axis direction, and $\eta_k=(r_k,L_k,s_k)\in\mathbb{R}^3$ contains the cylinder radius, length, and normalised axial position of the segment centre. Lastly, the PAD node is $x_{v,t}=\mathbf{0}_{22}$. This descriptor is used to study contact-aware control and graph-based information routing; estimating such quantities from noisy physical sensors is outside the scope of this simulation study.

To represent extended obstacles, each cylindrical obstacle is discretised along its axis into bins of length $\ell_{\mathrm{seg}}$.
A bin is discovered when any agent comes within a sensing radius $\rho$ of that bin. The newly discovered bins are assigned to the first PAD slot available and then persist for the rest of the episode. The count of new bins at time $t$ is $N_t^{\mathrm{new}}$, which directly feeds the discovery bonus in the reward function.

The algorithm stores a dense incoming adjacency matrix $A_t\in\{0,1\}^{|V|\times|V|}$, where $(A_t)_{u,v}=1$ means that a message is allowed from the source node $v$ to the target node $u$ (i.e., $v\rightarrow u$). To encode the arm’s kinematic chain, the neighbouring actuators are connected bidirectionally: $(A_t)_{i,i+1}=(A_t)_{i+1,i}=1$ for $i=1,\dots,N-1$. Finally, proximity sensing is modelled by directed entity$\rightarrow$agent edges. An active obstacle node $k$ with centre $c_k$ and radius $r_k$ adds $(A_t)_{i,k}=1$ whenever $\|p_{i,t}-c_k\|_2-r_k<\rho$. Importantly, obstacle nodes do not receive messages from agents, they only broadcast to agents when sensed.

\subsection{Attention-based Message Passing}
Given the constructed graph $G_t$, SoftGM uses a two-stage graph attention network to compute a context-aware embedding for each agent. Firstly, each node feature is augmented with its type embedding, which is $\tilde x_{v,t}=[x_{v,t}\,\|\,e(\tau_v)]$, and project it to the hidden space $h^{(0)}_{v,t}=W_{\mathrm{in}}\tilde x_{v,t}$.
The algorithm then split message passing into two explicit stages:
(i) entity$\rightarrow$agent propagation to inject obstacle information into agents, and (ii) agent$\leftrightarrow$agent propagation to enable coordination. Using node-type indicator masks, two stage-specific adjacencies are formed:
\[
A_t^{E\rightarrow A}=A_t\odot(\mathbf{1}_A\mathbf{1}_E^\top)+I,
\qquad
A_t^{A\leftrightarrow A}=A_t\odot(\mathbf{1}_A\mathbf{1}_A^\top)+I,
\]
where $\mathbf{1}_A$ and $\mathbf{1}_E$ select the agent and entity nodes, respectively, and $I$ is the identity.

Within each stage, the algorithm applies $L$ layers of multi-head attention over the permitted neighbours. For a directed edge $j\rightarrow i$ and head $h\in\{1,\dots,H\}$, the features are linearly transformed as
$u_i^{(h)}=W^{(h)}h_i$ and $u_j^{(h)}=W^{(h)}h_j$. Then a renormalised compatibility score is calculated using the standard GAT mechanism \cite{b22},
\[
e_{ij}^{(h)}=\mathrm{LeakyReLU}\!\left( (a^{(h)})^\top [u_i^{(h)} \,\|\, u_j^{(h)}] \right).
\]
Normalisation in the incoming neighbours $\mathcal{N}(i)$ of node $i$ yields attention weights
$\alpha_{ij}^{(h)}=\mathrm{softmax}_{j\in\mathcal{N}(i)}(e_{ij}^{(h)})$.
The updated representation aggregates neighbour information as
\[
m_i^{(h)}=\sum_{j\in\mathcal{N}(i)} \alpha_{ij}^{(h)}\,u_j^{(h)}.
\]
Attention weights $\alpha_{ij}^{(h)}$ allow the policy to suppress irrelevant nodes such as distant or uninformative obstacle segments, and emphasise only the most relevant interactions related to the task. $g_{i,t}$ is taken as the final embedding of an agent node $i$ after the two-stage attention network.
These $g_{i,t}$ are concatenated with local observations inside the actor, while the critic uses the mean pooled summary $\bar g_t=\frac{1}{N}\sum_i g_{i,t}$ to estimate $V_\phi(s_t)$ under CTDE.

\section{Experiments}
All experiments are conducted in simulation using PyElastica, which implements the Cosserat-rod theory to simulate the continuous mechanics of the soft arm \cite{b40}. The soft robotic arm is simulated as a single Cosserat rod with $60$ elements, base length $1.0\,\mathrm{m}$ and base radius $0.05\,\mathrm{m}$. 
The material parameters are set to density $\rho=1000.0\,\mathrm{kg\,m^{-3}}$, Young's modulus $E=5\times 10^{5}\,\mathrm{Pa}$, and shear modulus $G=2\times 10^{5}\,\mathrm{Pa}$. We use an analytical linear damper with damping constant $0.3$. The contact is modelled with stiffness $k=5\times10^{4}\,\mathrm{N\,m^{-1}}$ and damping $\nu=50.0\,\mathrm{N\,s\,m^{-1}}$. The simulator was wrapped in an OpenAI Gym interface to enable on-policy rollouts and parallel evaluation under identical control and training conditions \cite{b41}.

\subsection{Simulation Setup}
The experiment evaluates three environment instances that share the same Cosserat-rod soft-arm model and goal-reaching objective, but with the presence of increasingly complex external obstacles.
\begin{itemize}
    \item \textbf{Basic (no obstacle).} The workspace contains only the soft arm and the goal. This setting limits the learning difficulty to high-dimensional, coupled soft-body dynamics without additional constraints from external contact.
    \item \textbf{Structured obstacles.} Two fixed rigid cylindrical rods are placed between the arm and the goal region. This requires the controller to coordinate the arm while avoiding contacts.
    \item \textbf{Wall-with-hole.} A rigid wall is formed by an array of fixed cylindrical elements arranged to create a planar barrier with an opening. The target lies behind the wall, so the arm must pass through the opening to succeed. This setting induces intensive exploration because the passage location cannot be inferred from free-space kinematics alone and must be discovered through interaction during execution.
\end{itemize}

\begin{figure}[!t]
    \centering
    \includegraphics[width=\linewidth, keepaspectratio]{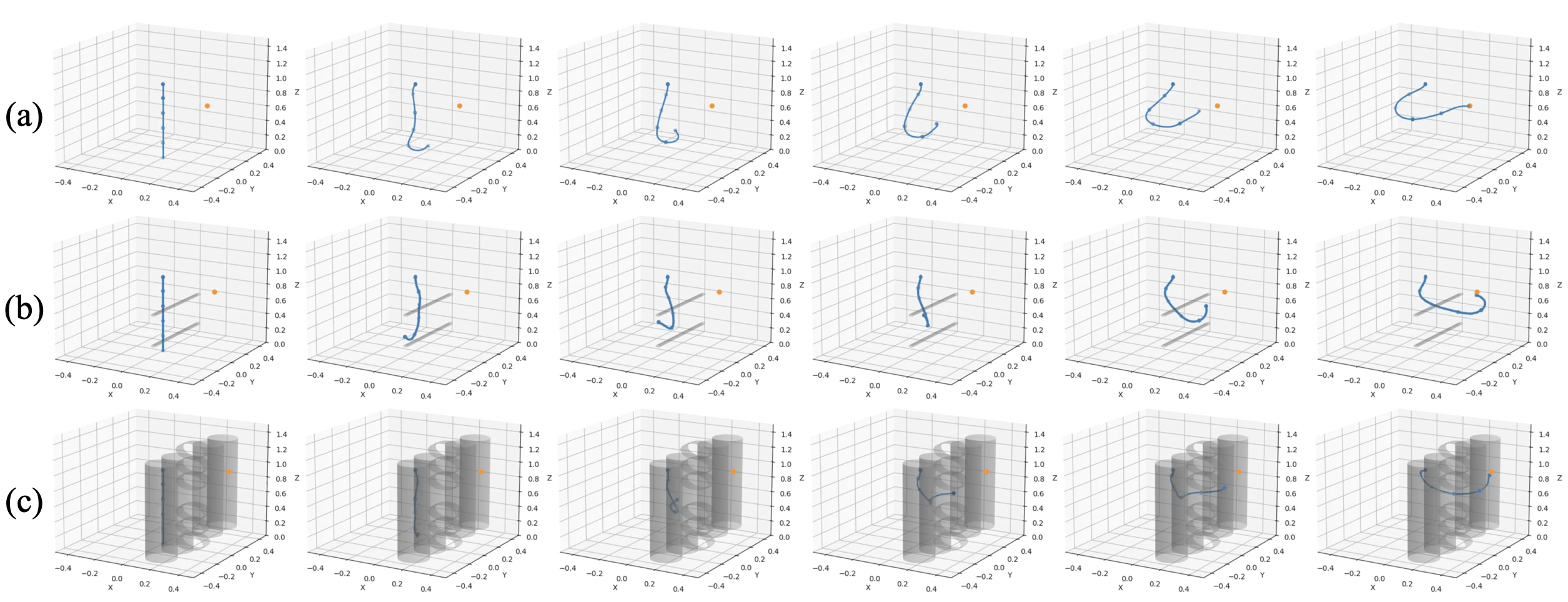}
    \caption{Snapshots of the simulated soft robotic arm (blue) in the three task scenarios. (a) Basic (no obstacle). (b) Structured obstacles. (c) Wall-with-hole.}
    \label{fig2} 
    \vspace{-1.5em} 
\end{figure}

The three testing scenarios are shown in Figure \ref{fig2}. The task of the arm in every episode is to move its end-effector (tip) to a randomly sampled target position $x^\star\in\mathbb{R}^3$. We denote the end-effector position by $x_t^{\mathrm{tip}}$ and define the tip-to-target distance as $d_t=\|x_t^{\mathrm{tip}}-x^\star\|_2$. An episode is considered successful when $d_t<r_{\mathrm{succ}}$, where $r_{\mathrm{succ}}$ is a fixed success radius. Episodes terminate upon success or when the maximum horizon is reached. Unless otherwise stated, all experiments use $N=6$ actuator agents, maximum horizon $T_{\max}=2000$ control steps, success radius $r_{\mathrm{succ}}=0.05\,\mathrm{m}$, obstacle discretisation length $\ell_{\mathrm{seg}}=0.05\,\mathrm{m}$, stuck threshold $\varepsilon=10^{-4}$, torque limit $\tau_{\max}=25.0$, and obstacle sensing radius $\rho=0.15\,\mathrm{m}$ for obstacle environments. The number of active obstacle nodes varies with the discovered obstacle segments in each scenario. This randomised target-reaching protocol encourages policies that generalise beyond a single memorised goal and provides a consistent performance across the three obstacle configurations.

\subsection{Comparisons with Baselines}
\label{subsec:baselines}
\paragraph{Baseline algorithms}
To evaluate the benefits of SoftGM, we compare it against six widely used MARL actor-critic baselines that cover both independent learners (IDDPG, IPPO, and ISAC) and CTDE methods (MADDPG, MAPPO, and MASAC). To ensure a fair comparison, all baseline methods and SoftGM are trained and evaluated in the same simulator environments with identical termination conditions, reward computation, action bounds, and evaluation protocol. Most importantly, each method receives exactly the same information content. The per-agent observation vector provided by the environment is identical across algorithms. The only difference lies in how each method processes these observations, not in what is observed. All benchmarks are implemented and executed using the BenchMARL pipeline, which standardises rollout collection, training and evaluation scheduling, and logging, thereby minimising implementation-induced discrepancies across methods \cite{b42}. We focus on standard BenchMARL baselines to provide controlled comparisons under identical observation, reward, and training pipelines; comparisons with specialised agent-entity or modular attention methods are left for future work.

\paragraph{Evaluation metrics and rationale}
We evaluate each method using complementary metrics that jointly capture task completion, solution quality, and control efficiency:
\begin{itemize}
    \item \textbf{Episodic reward.} We report the discounted return $R=\sum_{t=0}^{T-1}\gamma^t r_t$, which directly reflects the training objective. 
    \item \textbf{Evaluation success rate.} We report the evaluation success rate $\mathrm{SR}=\frac{1}{M}\sum_{m=1}^M \mathbb{I}[d_T^{(m)}<r_{\mathrm{succ}}]$ over $M$ evaluation episodes, measuring the fraction of trials that reach the target within the success radius.
    \item \textbf{Mean episode length.} We report the mean episode length $\bar T=\frac{1}{M}\sum_{m=1}^M T_m$, where $T_m$ is the number of control steps taken before the termination of episode $m$. Shorter episodes typically indicate faster goal attainment, and thus provide a measure of reaching efficiency in addition to success.
    \item \textbf{Mean tip travel distance.} We measure motion efficiency by the total path length travelled by the soft rod's end-effector during an episode. Lower values indicate more direct and efficient reaching behaviour, while higher values reflect detours due to exploration, obstacle interactions, or oscillatory motions.
    \item \textbf{Mean torque magnitude.} Finally, to characterise the control effort and aggressiveness of actuation, we report the mean torque magnitude as the average of the agent time $\bar a=\frac{1}{M}\sum_{m=1}^M \frac{1}{T_m N}\sum_{t=0}^{T_m-1}\sum_{i=1}^N \|u_{t}^{i,(m)}\|_2$. This metric helps to assess control feasibility under actuator limits, as well as robustness in contact-rich interactions.
\end{itemize}
These metrics provide a balanced comparison that measures learning dynamics, task performance, and control effort.

\subsection{Results}

\begin{figure}[!t]
    \centering
    \includegraphics[width=\linewidth, keepaspectratio]{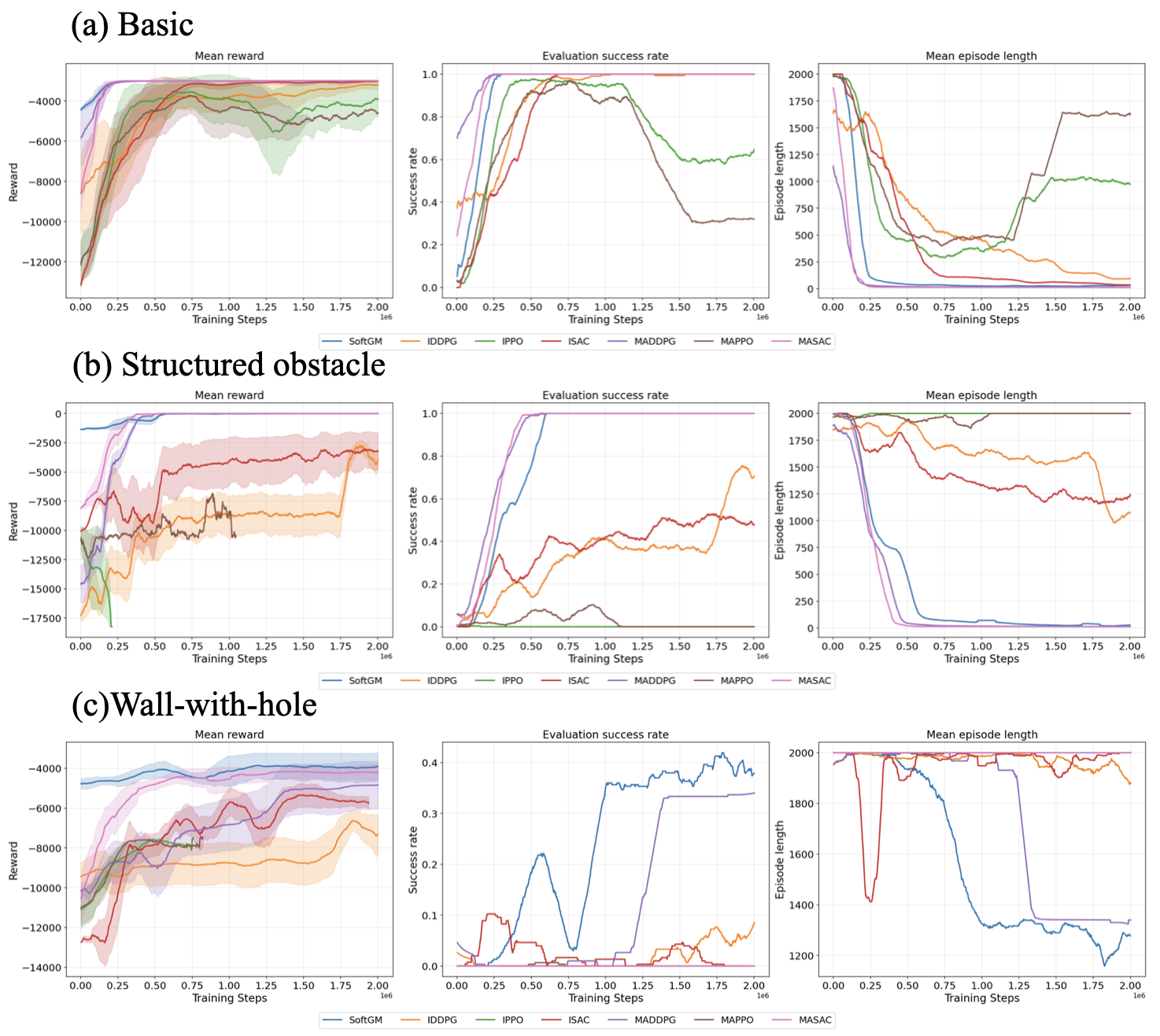}
    \caption{Learning curves of episodic rewards, evaluation success rates and mean episode lengths for SoftGM and the six MARL benchmarks in the three task scenarios over 3 random seeds.}
    \label{fig3} 
    \vspace{-1.5em} 
\end{figure}
For training, we used actor and critic learning rates of $1\times10^{-4}$ and $3\times10^{-4}$. The discount factor and GAE parameters are set to $\gamma=0.99$ and $\lambda_{\mathrm{GAE}}=0.95$. The clipping epsilon is $\epsilon=0.2$, with entropy coefficient $0.05$ and maximum gradient norm $0.5$. For the GAT network, we use hidden dimension 128 with 4 attention heads and 2 GNN layers, with a node-type embedding of dimension 8.

Figure \ref{fig3} reports the learning curves for the mean episodic reward, the evaluation success rate, and the mean episode length for SoftGM and six MARL baselines for the three scenarios of increasing contact complexity. All curves are averaged over three random seeds, with shaded regions indicating the standard error. In the obstacle-free task, most methods quickly achieve high success rates and short episodes once they learn stable actuation in the coupled soft-body dynamics. SoftGM converges rapidly and stably, matching the strong CTDE baselines (MADDPG/MASAC) in both success rate and episode length. The PPO-based methods exhibit notable late-training degradation, indicating that PPO-based optimisation can become unstable at later stages of training in highly coupled soft-body control. When structured obstacles are introduced, reliable coordination under contact becomes critical. SoftGM maintains near-perfect performance compared to strong CTDE baselines and has slightly shorter mean episode length. IDDPG and ISAC have much poorer performance, and PPO-based methods perform very poorly in this setting. PPO’s entropy-regularised, clipped updates can be hard to tune in contact-rich, long soft-body tasks; sustained exploration may destabilise learning, leading to performance collapse and, in some cases, extreme actions that break the simulation model. The wall-with-hole task magnifies the difference the most. SoftGM achieves the best overall performance among the tested methods, attaining the highest reward and success rate while reducing episode length substantially below the horizon, indicating more efficient searching behaviour. However, the absolute success rate remains modest, so this result should be interpreted as improved contact-guided exploration rather than a fully solved controller for highly constrained passages. MADDPG shows delayed partial improvement later in training, whereas the remaining baselines achieve near-zero success and remain at horizon-length episodes. Overall, these curves support that SoftGM's online obstacle discovery and graph-attention message passing enable it to focus on the few contact-relevant interactions at each moment, yielding better coordinated control in complex, contact-rich simulated environments.
\begin{figure*}[!t]
    \centering
    \includegraphics[width=\textwidth, keepaspectratio]{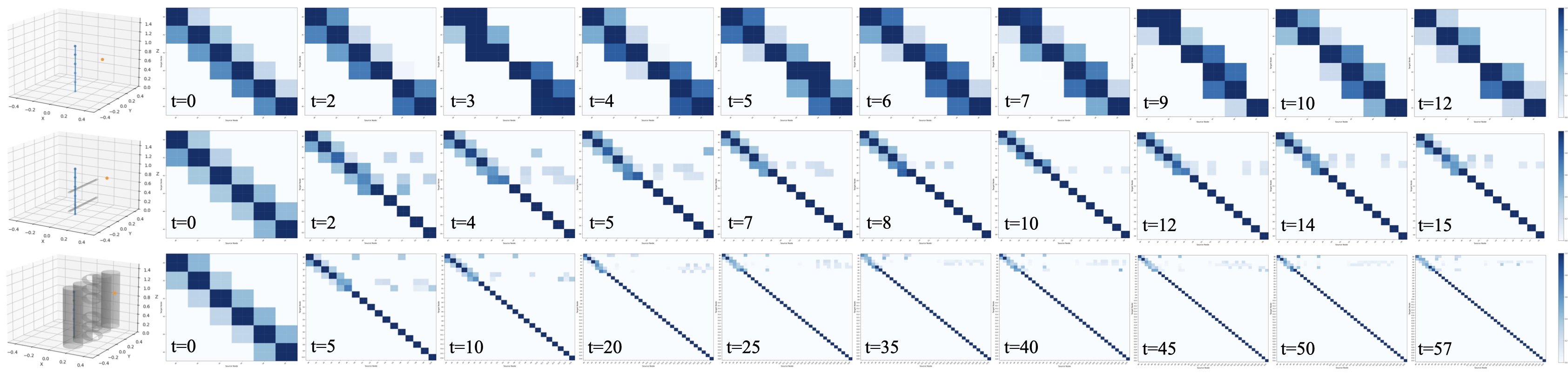}
    \caption{Visualisation of the attention matrices of the soft rod in three task scenarios. Each row shows the change of the attention matrix throughout one successful episode in the three scenarios. Darker colours indicate higher attention weights.}
    \label{fig4} 
    \vspace{-1em} 
\end{figure*}

\begin{table*}[t]
\centering
\footnotesize
\setlength{\tabcolsep}{4.0pt}
\renewcommand{\arraystretch}{1.25}
\caption{Evaluation results (mean $\pm$ std over 3 seeds $\times$ 100 episodes) across three scenarios. Tip travel distance denotes the total distance travelled by the end-effector (tip) during an episode. Mean torque magnitude is computed on the applied torques $u_t^i$. A dash indicates that the simulation became unstable or the policy failed before a meaningful travel-distance or torque statistic could be computed.}
\resizebox{\textwidth}{!}{%
\begin{tabular}{||l|c|c|c|c||c|c|c|c||c|c|c|c||}
\hline
& \multicolumn{4}{c||}{\textbf{Basic}} & \multicolumn{4}{c||}{\textbf{Structured Obstacles}} & \multicolumn{4}{c||}{\textbf{Wall-with-hole}} \\
\cline{2-13}
\textbf{Algorithm}
& \makecell{\textbf{Mean}\\\textbf{Ep. Len}}
& \makecell{\textbf{Mean Tip}\\\textbf{Travel Dist (m)}}
& \makecell{\textbf{Mean}\\\textbf{Torque Mag}}
& \makecell{\textbf{Success}\\\textbf{Rate}}
& \makecell{\textbf{Mean}\\\textbf{Ep. Len}}
& \makecell{\textbf{Mean Tip}\\\textbf{Travel Dist (m)}}
& \makecell{\textbf{Mean}\\\textbf{Torque Mag}}
& \makecell{\textbf{Success}\\\textbf{Rate}}
& \makecell{\textbf{Mean}\\\textbf{Ep. Len}}
& \makecell{\textbf{Mean Tip}\\\textbf{Travel Dist (m)}}
& \makecell{\textbf{Mean}\\\textbf{Torque Mag}}
& \makecell{\textbf{Success}\\\textbf{Rate}} \\
\hline

IDDPG
& $15.0 \pm 0.0$ & $1.0718 \pm 0.0013$ & $71.36 \pm 0.12$ & 100.00\%
& $904.1 \pm 558.5$ & $42.9560 \pm 30.9425$ & $72.53 \pm 1.83$ & 82.00\%
& $1913.8 \pm 306.8$ & $109.2686 \pm 32.5673$ & $81.33 \pm 6.05$ & 11.00\% \\
\hline

IPPO
& $831.9 \pm 884.5$ & $15.3576 \pm 24.1946$ & $71.71 \pm 10.48$ & 65.67\%

& $2000.0 \pm 0.0$ & \textemdash & \textemdash & 0.00\%
& $2000.0 \pm 0.0$ & \textemdash & \textemdash & 0.00\% \\
\hline

ISAC
& $18.4 \pm 1.8$ & $1.1765 \pm 0.0696$ & $60.49 \pm 0.87$ & 100.00\%
& $1372.1 \pm 886.4$ & $42.2383 \pm 29.7148$ & $62.19 \pm 1.10$ & 33.67\%
& $2000.0 \pm 0.0$ & $55.1215 \pm 10.7395$ & $58.00 \pm 2.59$ & 0.00\% \\
\hline

MADDPG
& $12.0 \pm 0.0$ & $1.0506 \pm 0.0006$ & $81.00 \pm 0.04$ & 100.00\%
& $12.3 \pm 0.5$ & $0.9763 \pm 0.0134$ & $80.08 \pm 0.68$ & 100.00\%
& $1487.6 \pm 865.4$ & $59.5494 \pm 40.2890$ & $79.66 \pm 3.94$ & 26.15\% \\
\hline

MAPPO
& $1512.3 \pm 725.0$ & $29.2250 \pm 22.2442$ & $78.21 \pm 0.24$ & 33.33\%
& $2000.0 \pm 0.0$ & \textemdash & \textemdash & 0.00\%
& $2000.0 \pm 0.0$ & \textemdash & \textemdash & 0.00\% \\
\hline

MASAC
& $12.9 \pm 0.2$ & $1.0605 \pm 0.0143$ & $66.79 \pm 1.21$ & 100.00\%
& $12.5 \pm 0.6$ & $0.9444 \pm 0.0201$ & $63.59 \pm 1.48$ & 100.00\%
& $2000.0 \pm 0.0$ & $19.5661 \pm 11.0627$ & $64.13 \pm 3.52$ & 0.00\% \\
\hline

\textbf{SoftGM}
& $\bm{17.1 \pm 17.2}$ & $\bm{1.2555 \pm 0.8521}$ & $\bm{80.01 \pm 1.59}$ & \textbf{100.00\%}
& $\bm{16.1 \pm 2.9}$ & $\bm{1.1161 \pm 0.1489}$ & $\bm{76.17 \pm 1.40}$ & \textbf{100.00\%}
& $\bm{1297.5 \pm 907.0}$ & $\bm{66.4659 \pm 46.5124}$ & $\bm{72.24 \pm 1.64}$ & \textbf{41.33\%} \\
\hline
\end{tabular}%
}
\label{tab:eval_results_compact}
\vspace{-1em} 
\end{table*}

\begin{table*}[t]
\centering
\footnotesize
\setlength{\tabcolsep}{4.0pt}
\renewcommand{\arraystretch}{1.25}
\caption{SoftGM robustness evaluation (mean $\pm$ std over 3 seeds $\times$ 100 episodes) under three non-ideal conditions in scenario (c).}
\resizebox{\textwidth}{!}{%
\begin{tabular}{||l|c|c|c|c||c|c|c|c||c|c|c|c||}
\hline
& \multicolumn{4}{c||}{\textbf{Noise}} & \multicolumn{4}{c||}{\textbf{One-section Failure}} & \multicolumn{4}{c||}{\textbf{Disturbance}} \\
\cline{2-13}
\textbf{Algorithm}
& \makecell{\textbf{Mean}\\\textbf{Ep. Len}}
& \makecell{\textbf{Mean Tip}\\\textbf{Travel Dist (m)}}
& \makecell{\textbf{Mean}\\\textbf{Torque Mag}}
& \makecell{\textbf{Success}\\\textbf{Rate}}
& \makecell{\textbf{Mean}\\\textbf{Ep. Len}}
& \makecell{\textbf{Mean Tip}\\\textbf{Travel Dist (m)}}
& \makecell{\textbf{Mean}\\\textbf{Torque Mag}}
& \makecell{\textbf{Success}\\\textbf{Rate}}
& \makecell{\textbf{Mean}\\\textbf{Ep. Len}}
& \makecell{\textbf{Mean Tip}\\\textbf{Travel Dist (m)}}
& \makecell{\textbf{Mean}\\\textbf{Torque Mag}}
& \makecell{\textbf{Success}\\\textbf{Rate}} \\
\hline

SoftGM
& $1312.7 \pm 905.4$ & $66.9108 \pm 46.3512$ & $72.33 \pm 1.61$ & 37.33\%
& $1464.5 \pm 772.0$ & $68.6340 \pm 39.0639$ & $73.17 \pm 0.63$ & 36.00\%
& $1334.5 \pm 857.7$ & $66.6386 \pm 43.0307$ & $72.12 \pm 1.22$ & 40.33\% \\
\hline

\end{tabular}%
}
\label{tab:softgm_robustness}
\vspace{-1em} 
\end{table*}

\section{Evaluation and Discussion}


\subsection{Comparison against MARL baselines}
Table~I summarises the evaluation performance of SoftGM against six MARL baselines across three scenarios of increasing contact complexity. We compare success rate, mean episode length, tip travel distance, and mean torque magnitude.

In the basic scenario, most methods achieve high success, showing that the task mainly tests stable actuation in coupled soft-body dynamics. In the structured-obstacle scenario, SoftGM maintains $100\%$ success and remains competitive with the strongest CTDE baselines, while independent and PPO-based methods degrade substantially. The wall-with-hole task is the most challenging because success requires contact-guided exploration to locate and traverse the opening. In this setting, SoftGM achieves the highest success rate among all tested methods ($41.33\%$), outperforming MADDPG ($26.15\%$) and IDDPG ($11.00\%$), while all remaining baselines fail to solve the task. Although the absolute success rate remains modest, the results show that online obstacle discovery and attention-based message passing improve contact-relevant information routing in the most constrained environment.

\subsection{Qualitative Visualization}
Figure \ref{fig4} visualises the attention matrices of SoftGM across the three scenarios. In the obstacle-free case, attention remains broadly distributed around neighbouring sections, reflecting the need for local coordination in coupled soft-body dynamics. With structured obstacles, attention becomes more selective and shifts toward relevant obstacle segments as contact interactions emerge. In the wall-with-hole task, the attention pattern becomes highly time-varying: different arm sections attend to different wall regions as the rod searches for and passes through the opening. These results support the intended role of attention in suppressing irrelevant interactions while prioritising contact-relevant information.

\subsection{Robustness evaluation}
To assess robustness beyond idealised simulation conditions, we evaluate SoftGM under observation noise, one-section actuation failure, and transient external disturbance.

\begin{itemize}
  \item \textbf{Noisy observations.} Zero-mean Gaussian noise is added independently to selected observation channels: position noise with standard deviation $0.1$, velocity noise with standard deviation $0.1$, and torque-related observation noise with standard deviation $0.2$.
  \item \textbf{One-section failure.} The actuator (agent) with index $2$ is forced to output zero actuation throughout the episode. This section does not exert any torque regardless of the policy output, emulating a hard loss of actuation at one segment.
  \item \textbf{Disturbance.} A transient external force of $15\,\mathrm{N}$ is applied to $6$ consecutive frames starting at step $5$, representing an impulsive disturbance during early motion.
\end{itemize}

This robustness evaluation probes the sensitivity to observation corruption, partial loss of actuation authority, and impulsive external perturbations in contact-rich soft robotic arm operation.

Table~II reports robustness results over 3 seeds $\times$ 100 episodes. Under observation noise, one-section failure, and transient disturbance, SoftGM maintains success rates of 37.33\%, 36.00\%, and 40.33\%, respectively, which are comparable to the nominal wall-with-hole performance. The episode lengths and tip travel distances remain within a similar range across all three conditions, while the mean torque magnitude stays stable. These results suggest that SoftGM's distributed message passing can partially compensate for corrupted observations, local actuation loss, and short external perturbations without relying on overly aggressive control.

\subsection{Ablation study: Two-stage Attention-based Message Passing}
\begin{figure}[!t]
    \centering
    \includegraphics[width=\linewidth, keepaspectratio]{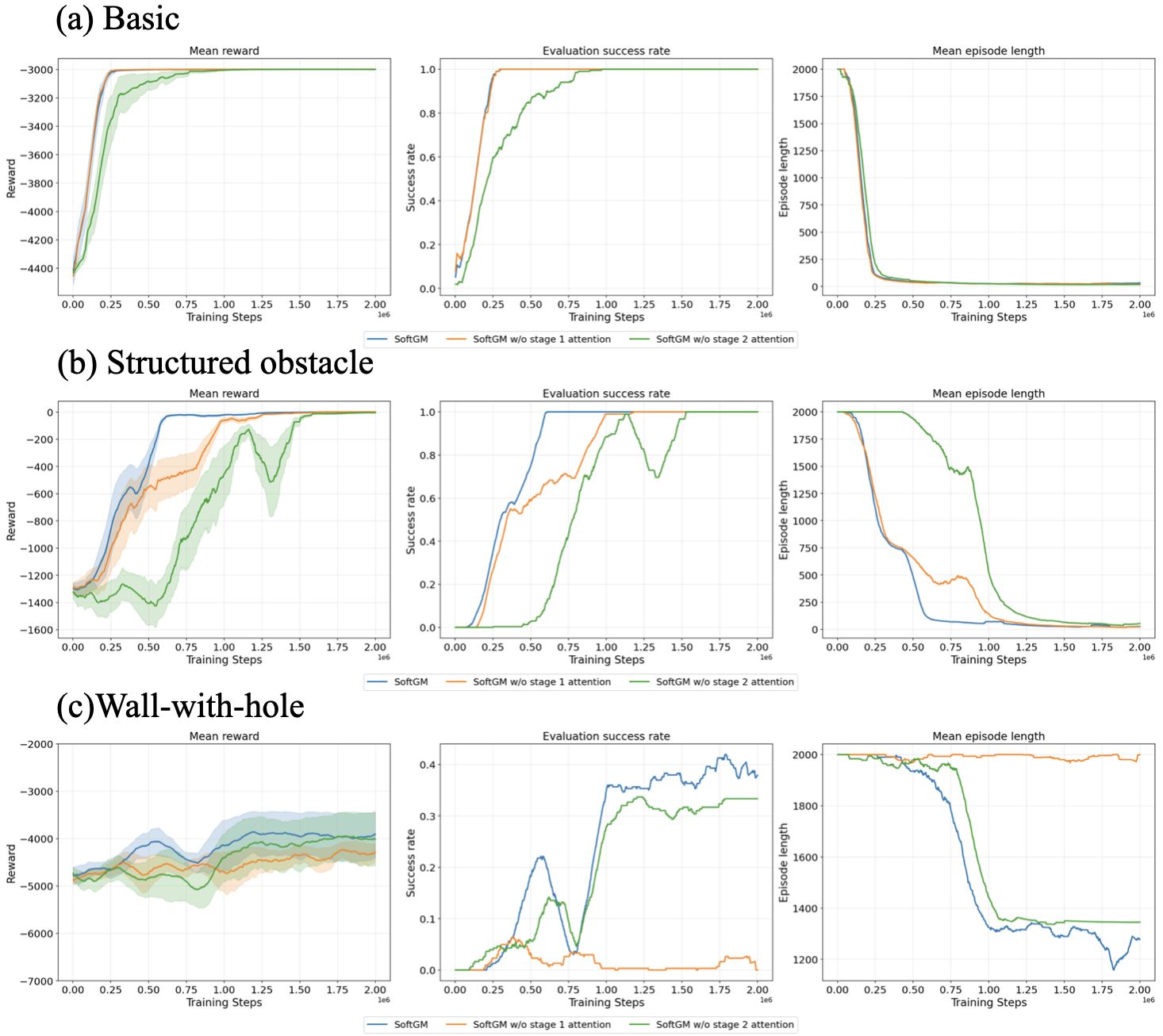}
    \caption{The comparison of SoftGM with its ablations (SoftGM without stage 1 attention and SoftGM without stage 2 attention) in the three task scenarios.}
    \label{fig6} 
    \vspace{-1.5em} 
\end{figure}

Figure \ref{fig6} compares SoftGM with two ablations that remove stage 1 attention (entity$\rightarrow$agent) or stage 2 attention (agent$\leftrightarrow$agent), while keeping the remaining components unchanged. In scenario (a), all three variants quickly converge to $100\%$ success rate with short episode lengths. Removing stage-1 attention has a negligible effect because there are no obstacle entities to attend to. Removing stage-2 attention slightly slows convergence, indicating that explicit agent-to-agent coordination is still beneficial even in obstacle-free reaching. In scenario (b), the full SoftGM converges to perfect success in short episodes earlier than both ablations. Removing stage 2 attention notably slows down learning and introduces transient instability, suggesting that agent-to-agent attention is important for propagating locally sensed interaction cues along the arm to maintain coordinated, contact-aware motion. Scenario (c) highlights that stage 1 attention is essential for exploration in complex environments, without which success remains near zero. Overall, the ablation study supports that stage 1 attention is the key mechanism for environment discovery, while stage-2 attention supports stable distributed coordination in complex environments.

\subsection{Limitations}
The present results should be interpreted within three limitations. First, all evaluations are simulation-based; therefore, the results demonstrate algorithmic feasibility in PyElastica rather than full hardware readiness. Second, although SoftGM achieves the best success rate in the wall-with-hole task, the absolute success rate remains modest and the variance is high. Third, discovered obstacle nodes use simplified geometric descriptors available in the simulator, which are suitable for controlled studies of contact-aware control and message routing but shows limitations in arbitrary real-world geometry reconstruction.

\section{Conclusion}
This paper presented SoftGM, an octopus-inspired GNN-based distributed control architecture for Cosserat-rod soft robotic arms in contact-rich simulated environments. By combining online obstacle discovery with two-stage graph attention, SoftGM selectively routes contact-relevant information while preserving coordination among neighbouring arm sections. Across three simulated tasks, SoftGM matches strong CTDE baselines in simpler settings and achieves the best performance in the wall-with-hole task, where exploration and selective information routing are critical.

Future work will focus on sim-to-real transfer, adaptive graph construction for dynamic obstacles, comparisons with specialised graph-attention MARL baselines, and generalisation across different segment counts and modular arm configurations.

\section*{Acknowledgment}
Some sentences in the paper have been paraphrased and improved using AI tools.

\end{document}